\documentclass[letterpaper, 10 pt, conference]{ieeeconf}  
\usepackage{graphicx}
\usepackage{booktabs}
\usepackage{stfloats}
\usepackage[justification=centering]{caption}
\usepackage{cite}

\IEEEoverridecommandlockouts                              

\title{\LARGE \bf
An Ensemble Learning Framework for Vehicle Trajectory Prediction in Interactive Scenarios*
}

\author{Zirui Li$^{1, 2, 3}$, Yunlong Lin$^{1, 2}$, Cheng Gong$^{2}$, \textit{Student Member IEEE},\\ 
Xinwei Wang$^{3}$,  Qi Liu$^{2}$, Jianwei Gong$^{2}$, \textit{Member IEEE}, 
Chao Lu$^{2}$, \textit{Member IEEE}
\thanks{$^{1}$Zirui Li and Yunlong Lin contribute equally.}
\thanks{*This work was supported by the National Natural Science Foundation of China under Grants 61703041 and U19A2083.}
\thanks{$^{2}$Zirui Li, Yunlong Lin, Cheng Gong, Qi Liu, Jianwei Gong and Chao Lu are with the School of Mechanical Engineering, Beijing Institute of Technology, Beijing 100081, China.(E-mails: {\tt\small ziruili.work.bit@gmail.com}; {\tt\small 1120181526@bit.edu.cn}; {\tt\small chenggong@bit.edu.cn}; {\tt\small 3120195257@bit.edu.cn}; {\tt\small gongjianwei@bit.edu.cn}; {\tt\small chaolu@bit.edu.cn})}%
\thanks{$^{3}$Zirui Li and Xinwei Wang are with the Department of Transport and Planning, Faculty of Civil Engineering and Geosciences, Delft University of Technology, Stevinweg 1, 2628 CN Delft, The Netherlands(E-mails: {\tt\small x.w.wang@tudelft.nl})}.
\thanks{(Corresponding author: J. Gong)}%
}

\begin{document}

\maketitle
\thispagestyle{empty}
\pagestyle{empty}

\begin{abstract}

Precisely modeling interactions and accurately predicting trajectories of surrounding vehicles are essential to the decision-making and path-planning of intelligent vehicles. This paper proposes a novel framework based on ensemble learning to improve the performance of trajectory predictions in interactive scenarios. The framework is termed Interactive Ensemble Trajectory Predictor (IETP). IETP assembles interaction-aware trajectory predictors as base learners to build an ensemble learner. Firstly, each base learner in IETP observes historical trajectories of vehicles in the scene. Then each base learner handles interactions between vehicles to predict trajectories. Finally, an ensemble learner is built to predict trajectories by applying two ensemble strategies on the predictions from all base learners. Predictions generated by the ensemble learner are final outputs of IETP. In this study, three experiments using different data are conducted based on the NGSIM dataset. Experimental results show that IETP improves the predicting accuracy and decreases the variance of errors compared to base learners. In addition, IETP exceeds baseline models with 50\% of the training data, indicating that IETP is data-efficient. Moreover, the  implementation of IETP is publicly available at https://github.com/BIT-Jack/IETP.

\end{abstract}

\section{Introduction}

Developing intelligent vehicles with socially compliant and conventional driving behaviors is significant to traffic safety and road mobility \cite{li2022personalized,lu2019transfer,lu2019virtual}. In interactive scenarios such as urban roads and freeways, the dynamic motions of surrounding vehicles limit the availability of some driving actions. Thus, intelligent vehicles need to predict future trajectories of surrounding vehicles before making decisions, and planning paths \cite{deo2018would}. Accurately predicting the future trajectories of surrounding vehicles becomes a fundamental ability of intelligent vehicles. However, predicting trajectories in interactive scenarios is challenging due to the complexity and uncertainty of interactions between vehicles. For example,  various drivers' driving styles and different destinations of vehicles reveal the complexity. The randomness of driving behaviors indicates the uncertainty \cite{li2020importance}. Therefore, how to precisely model interactions between vehicles becomes a crucial problem of predicting vehicle trajectories in interactive scenarios.

Compared to traditional trajectory predicting methods such as Kalman filter \cite{ess2010object} and social force model \cite{luber2010people}, deep learning-based methods have shown outstanding performance to model interactions between traffic agents such as vehicles and pedestrians \cite{liu2021survey}. In deep learning, a type of neural network termed Long Short-Term Memory (LSTM) has been shown to successfully handle tasks with sequential inputs and outputs. Since trajectory predictions can be viewed as such sequential to sequential tasks, many LSTM-based methods are proposed for trajectory predictions in interactive scenarios.

LSTM-based methods \cite{alahi2016social, deo2018convolutional , xu2018collision} utilize the pooling mechanism to model social interactions. In \cite{alahi2016social}, social pooling layers are proposed to capture interactions among pedestrians. The spatial information of pedestrians is preserved through grid-based pooling. \cite{deo2018convolutional} extends the Social-LSTM \cite{alahi2016social} by applying convolutional layers to replace fully connected layers, which is proposed as a remedy to address generalization problems. \cite{zhang2019sr,zhao2020spatial,peng2021sra} adopt the attention mechanism to deal with interactive information. \cite{zhang2019sr} proposes a state refinement module for the LSTM network. Trajectories are predicted by utilizing the current intention of surrounding agents. In \cite{peng2021sra}, a Social Relationship Attention LSTM (SRA-LSTM) is proposed to predict future trajectories. Social relationship attention to aggregate movement information from neighbor agents is utilized to model the interactions in SRA-LSTM. These LSTM-based methods focus on designing special mechanisms to model interactions. The mainly used networks are LSTM. Furthermore, more types of networks such as Generative Adversarial Networks (GAN), Graph-based networks, and Temporal Convolutional Networks (TCN) are utilized to improve the modeling and predicting performance.

In \cite{gupta2018social}, tools from sequence prediction and generative adversarial networks are combined to predict socially plausible trajectories. \cite{kosaraju2019social} improves the Social-GAN model by introducing a flexible graph attention network. \cite{sun2020recursive} extracts interactive information into social behavior graphs. The graph convolutional neural network is then applied to propagate social interaction information in such graphs. \cite{zhao2020gisnet} proposes a graph-based information-sharing network (GISNet) to improve the accuracy of vehicle trajectory prediction compared to baselines in experiments. \cite{wang2021graphtcn} proposes a graph-based temporal convolutional network. Both the accuracy and efficiency of trajectory predictions are improved compared to baselines in experiments. \cite{li2021hierarchical} proposes a hierarchical Graph Neural Network (GNN) framework combined with LSTM to model interactions of heterogeneous traffic participants and predict their trajectories. Similarly, in \cite{li2021interactive}, by using GNN, interactions and trajectories are firstly modeled as spatial-temporal graphs. Then a GNN-based multitask learning framework is proposed to accurately predict trajectories of vehicles and pedestrians.

However, these previous studies only focus on the design of network structures to model interactions, and each model is used individually to predict trajectories. The combination of different models is ignored. Using an individual model to make predictions may lead to low predicting accuracy and the model can be sensitive to training data \cite{xing2020ensemble}. 

As discussed in \cite{sagi2018ensemble} and \cite{dong2020survey}, the individual model can be weak to make predictions. In comparison, ensemble learning methods that assemble individual models to make predictions can improve the predictive performance of an individual model \cite{dietterich2002ensemble}. For example, \cite{xing2020ensemble} proposes an ensemble learning model which improves the predicting accuracy for driver lane change intention inference. In \cite{xing2020ensemble}, a data augmentation scheme is firstly designed to increase the data volume and generate multiple training sets. Then, based on the Bootstrap aggregating method, different RNN models are assembled to inference the lane change intention. Results show that the accuracy and robustness of the RNN models for intention inference are improved by applying the ensemble method. To the best of our knowledge, there are few ensemble learning studies focusing on vehicle trajectory prediction. Therefore, this paper proposes a novel ensemble learning framework for vehicle trajectory prediction in interactive scenarios, and we term it as Interactive Ensemble Trajectory Predictor (IETP). The main contributions of this paper are as follows:
\begin{itemize}
\item A  novel ensemble learning framework is proposed for vehicle trajectory prediction in interactive scenarios. The framework assembles interaction-aware trajectory predictors as base learners to build an ensemble learner, improving the predicting performance of base learners.
\item Two ensemble learning strategies are proposed for trajectory predictors to handle the maneuvers classification and trajectory prediction tasks, respectively.
\item Three experiments using a different number of the data are conducted to evaluate the proposed framework. Moreover, a comparative study in the time-cost of the proposed framework is also presented.
\end{itemize}

The remainder of this paper is organized as follows. Section \ref{methods} introduces the construction of IETP. Section \ref{exp} shows experimental results and analysis. Lastly, the conclusion of this paper is presented in Section \ref{conclu}.

\section{Interactive Ensemble Trajectory Predictor}
\label{methods}
\begin{figure*}[ht]
      \centering
      \includegraphics[scale=1.0]{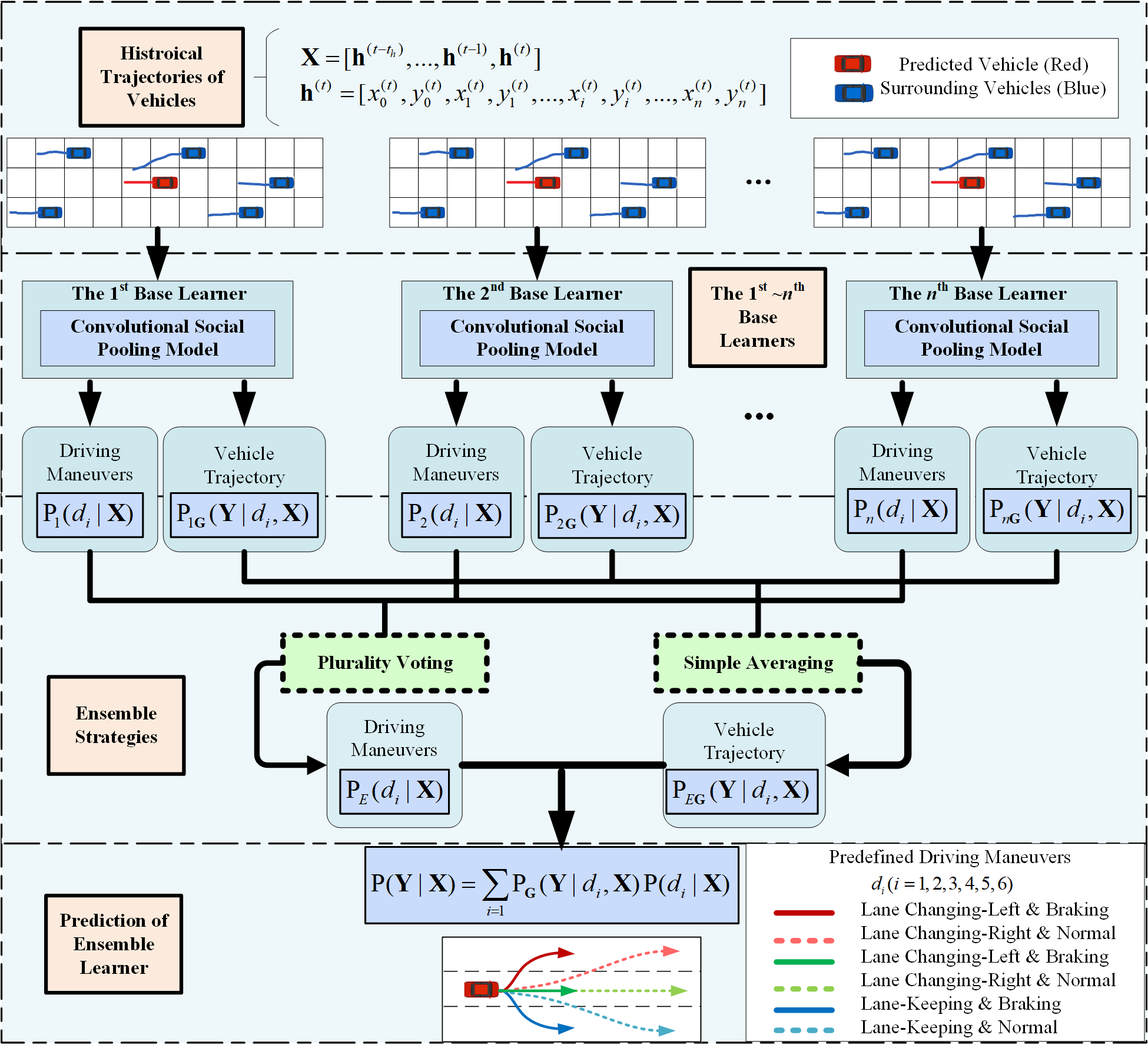}
      \captionsetup{font={small}}
      \caption{IETP assembles $n$ interaction-aware models as base learners. To demonstrate the performance of IETP, this study adopts Convolutional Social Pooling models as base learners in experiments. First, IETP assembles $n$ base learners by Bootstrap aggregating. Then, two specific ensemble strategies including plurality voting and simple averaging are used to build the ensemble learner.}
      \label{model}
   \end{figure*}

This section will introduce the proposed IETP, which is displayed in Fig. \ref{model}. IETP assembles interaction-aware models which consider interactions between vehicles as base learners to build an ensemble learner, predicting trajectories in interactive scenarios. Firstly, inputs of IETP are fed into each base learner. Each base learner then predicts trajectories based on handling interactions between vehicles. Finally, IETP obtains outputs by applying ensemble strategies on predictions from all base learners. IETP aims to provide a general ensemble learning approach for trajectory prediction in interactive scenarios. Convolutional Social Pooling models proposed in \cite{deo2018convolutional} are adopted as base learners in this study. It can be regarded as an example of IETP instead of a fixed approach.

\subsection{Problem Formulation}
\label{formulation}
According to \cite{deo2018convolutional}, the trajectory prediction is formulated as estimating the probability distribution of future positions of the predicted vehicle in this study. Inputs to IETP are historical trajectories of the predicted vehicle and surrounding vehicles. The historical trajectories can be described as:
\begin{equation}
\label{input_x}
        {\bf{X}} = [{{\bf{h}}^{(t - {t_h})}},...,{{\bf{h}}^{(t - 1)}},{{\bf{h}}^{(t)}}],
\end{equation}where $
{{\bf{h}}^{(t)}} = [x_0^{(t)},y_0^{(t)},x_1^{(t)},y_1^{(t)},...,x_i^{(t)},y_i^{(t)},...,x_n^{(t)},y_n^{(t)}]$ are x and y coordinates at time $t$. Here, the y-axis points to the driving direction. The x-axis points to the lateral direction. In vector ${{\bf{h}}^{(t)}}$, $x_0^{(t)}$ and $y_0^{(t)}$ are coordinates of the vehicle being predicted at time $t$, while $x_i^{(t)}$, $y_i^{(t)}$ ($i$=1,2,3,…,$n$) represent the coordinates of different surrounding vehicles at time $t$.
The output of IETP is a probability distribution over future coordinates of the predicted vehicle. If future coordinates are described as ${\bf{Y}}$, the output distribution can be described as:
\begin{equation}
    {\mathop{\rm P}\nolimits} ({\bf{Y}}|{\bf{X}}) = \sum\limits_{i = 1}^{6} {{{\mathop{\rm P}\nolimits} _{\bf{G}}}({\bf{Y}}|{d_i},{\bf{X}}){\mathop{\rm P}\nolimits} ({d_i}|{\bf{X}})}\label{output_p},
\end{equation}
where ${\bf{G}} = [{{\bf{G}}^{(t + 1)}},...,{{\bf{G}}^{(t + {t_f})}}]$ are parameters of a bivariate Gaussian distribution at each time step in the prediction horizon. Detailed parameters of a bivariate can be described as:
\begin{equation}
    {{\bf{G}}^{(t)}} = [m_x^{(t)},m_y^{(t)},s_x^{(t)},s_y^{(t)},{r^{(t)}}]\label{Gaussian},
\end{equation}
where $m_x^{(t)}$ and $m_y^{(t)}$ are mathematical expectations of the predicted future locations in the lateral and longitudinal direction at time $t$. While $s_x^{(t)}$ and $s_y^{(t)}$  are variances of ${\bf{X}}$ and ${\bf{Y}}$, and ${r^{(t)}}$ is the coefficient of association. ${d_i}$ ($i$=1, 2, 3, 4, 5, 6) represents six driving maneuvers defined in \cite{deo2018convolutional}. Detailed types of maneuvers are also shown in Fig. \ref{model}.

\subsection{Ensemble Learning Approaches for IETP}
\begin{figure}[t]
      \centering
      \includegraphics[scale=1.0]{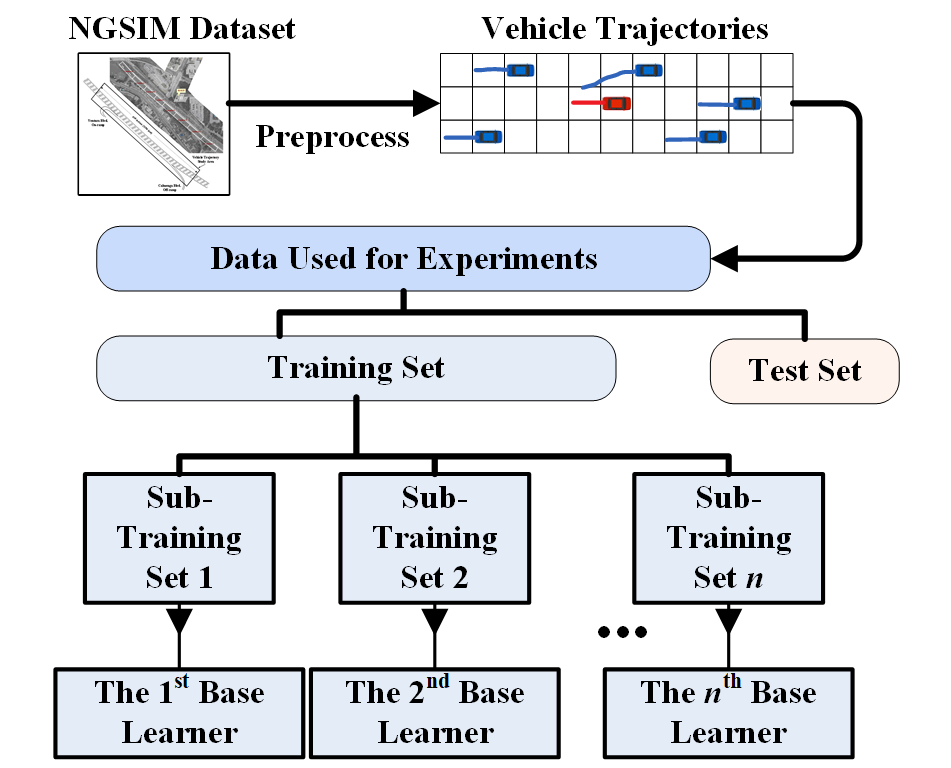}
      \captionsetup{font={small}}
      \caption{Data-processing in experiments.}
      \label{data_process}
   \end{figure}

Since the diversity of base learners is critical to the performance of the ensemble learner \cite{dong2020survey}, this study applies the Bootstrap aggregating (bagging) method to obtain diverse base learners. Firstly, the NGSIM dataset \cite{ngsimdata} is processed as shown in Fig. \ref{data_process}. The processed dataset consists of trajectories information. Secondly, several sub-training sets are obtained by random sampling with replacement based on the processed dataset. In detail, every sub-training set has the same number of samples as the whole training set. This type of sampling is named Bootstrap sampling \cite{efron1994introduction}. Then, base learners are trained on these sub-training sets, respectively. It should be noted that the trained weights of networks in these base learners are different due to diverse training samples. To distinguish different base learners, all base learners are numbered.

After obtaining diverse base learners, as shown in Fig. \ref{model}, inputs described as (\ref{input_x}) are fed into each base learner. Then, each base learner outputs prediction as (\ref{output_p}). Finally, IETP makes predictions by assembling predictions from base learners through ensemble strategies including plurality voting and simple averaging.

Plurality voting is a commonly used ensemble strategy in classification task \cite{sagi2018ensemble}. If labels in a classification task are defined as:

\begin{equation}
    C = \{ {c_1},{c_2},...,{c_j},...,{c_{\rm{T}}}\},\label{class_set}
\end{equation}
the subscript $j$ ($j$ = 1, 2, 3, …, T) represents the class of label, while T is the total number of labels. Each base learner will predict from set $C$. Thus, the prediction of each base learner on a sample ${\bf{x}}$, can be expressed as:
\begin{equation}
    {{\mathop{\rm B}\nolimits} _i}({\bf{x}}) = \{ {\mathop{\rm b}\nolimits} _i^1({\bf{x}}),{\mathop{\rm b}\nolimits} _i^2({\bf{x}}),...,{\mathop{\rm b}\nolimits} _i^j({\bf{x}}),...,{\mathop{\rm b}\nolimits} _i^{\rm{T}}({\bf{x}})\}.\label{pred_base}
\end{equation}
In (\ref{pred_base}),
\begin{equation}
    {\mathop{\rm b}\nolimits} _i^j({\bf{x}}) = \left\{ \begin{array}{l}
1{\rm{, predicted}}\\
0{\rm{, otherwise}}
\end{array} \right.
\end{equation}
   is the prediction made by the base learner ${{\mathop{\rm b}\nolimits} _i}$ on label ${c_j}$. Moreover, on every sample ${\bf{x}}$, only a label is predicted. Thus, the prediction made by each base learner can be seen as a one-hot vector. If $n$ base learners are used to build an ensemble learner, the predicted result of the ensemble learner can be described as:
\begin{equation}
\left\{ \begin{array}{l}
{{\mathop{\rm c}\nolimits} _{\rm{E}}}({\bf{x}}) = {c_j}\\
j = \mathop {\arg \max }\limits_j \sum\limits_{i = 1}^n {{\mathop{\rm b}\nolimits} _i^j({\bf{x}})} 
\end{array} \right.,
\end{equation}
where ${{\mathop{\rm c}\nolimits} _{\rm{E}}}({\bf{x}})$  is the prediction on sample ${\bf{x}}$ made by the ensemble learner. The result is the most voted label predicted by $n$ base learners. If more than one labels get the most votes, the result will be obtained from them randomly.

In IETP, each base learner makes trajectory prediction in terms of six driving maneuvers. Plurality voting method is applied to obtain ensemble driving maneuvers. It should be noted that, original predictions made by base learners are probability format, which can be expressed as:
\begin{equation}
    \left\{ \begin{array}{l}
[{p_1},{p_2},{p_3},{p_4},{p_5},{p_6}]\\
\sum\limits_{i = 1}^6 {{p_i} = 1} 
\end{array} \right.\label{pro_format},
\end{equation}
where ${p_i}$ ($i$ = 1, 2, 3, 4, 5, 6) is the probability value corresponding to a class of driving maneuver. Firstly, each original prediction is encoded into a one-hot vector as described in (\ref{pred_base}). Then, the plurality voting method is applied to these one-hot vectors to obtain ensemble learning results. Finally, the predicted one-hot vector is decoded back into the probability format as (\ref{pro_format}).

Simple averaging is a widely used ensemble strategy when dealing with regression tasks \cite{sagi2018ensemble}. Results from ensemble learners are obtained by averaging predicted values of all base learners. In this study, outputs of our proposed IETP are probability distributions as described in Section \ref{formulation}. Since the probability is assumed with a bi-variate Gaussian distribution, specific outputs can be described by five parameters of Gaussian distribution in (\ref{Gaussian}). Therefore, detailed outputs of IETP are averaged values of these parameters.

\subsection{Loss Fuction}
According to \cite{deo2018convolutional}, in the training process, the negative log-likelihood can be described as:
\begin{equation}
    - \log \left( {{{\mathop{\rm P}\nolimits} _{\bf{G}}}\left( {{\bf{Y}}|{d_{{\rm{true}}}},{\bf{X}}} \right){\mathop{\rm P}\nolimits} \left( {{d_{{\rm{true}}}}|{\bf{X}}} \right)} \right).\label{loss}
\end{equation}
Equation (\ref{loss}) is minimized overall training samples. In (\ref{loss}), ${d_{{\rm{true}}}}$ provided by each training instance, is the driving maneuver class that is actually performed.

\subsection{Implementation Details}
In this study, models observe historical trajectories for three seconds and predict trajectories for next five seconds. Moreover, each base learner in IETP is trained using the Adam \cite{kingma2014adam} optimizer with a learning rate of 0.001. According to \cite{deo2018convolutional}, the key parameters of each base learner is shown in Table \ref{keyparams}.  The leaky-ReLU activation with $\alpha$ equals 0.1 is used for all layers. The model is implemented using PyTorch\footnote{https://pytorch.org}.

\begin{table}[]
    \centering
    \captionsetup{font={small}}
    \caption{KEY PARAMETERS OF EACH BASE LEARNER IN IETP}
    \begin{tabular}{c c c c c}
    \toprule
    Order & Layers & Hidden states & Size & Depth \\ \midrule
    1     & LSTM encoder & 64 & - & - \\ 
    2     & Social tensor & - & 13x3 & - \\ 
    3     & Convolutional layer & - & 3x3 & 64 \\ 
    4     & Convolutional layer & - & 3x1 & 16 \\ 
    5     & Max pooling layer & - & 2x1 & - \\ 
    6     & LSTM decoder & 128 & - & - \\ 
    \bottomrule
    \end{tabular}
    \label{keyparams}
\end{table}

\section{Experiments}
\label{exp}

This section presents results and statistical analysis of experiments based on the NGSIM dataset. Three experiments using a different number of the data are conducted to evaluate IETP. In each experiment, 20 sub-training sets are obtained by the Bootstrap sampling. Then, these sub-training sets are used to train 20 base learners, respectively. Finally, 20 ensemble learners are built by assembling different base learners. Detailed experiments are introduced in the following.
 \begin{figure*}[bp]
      \centering
      \includegraphics[scale=1.0]{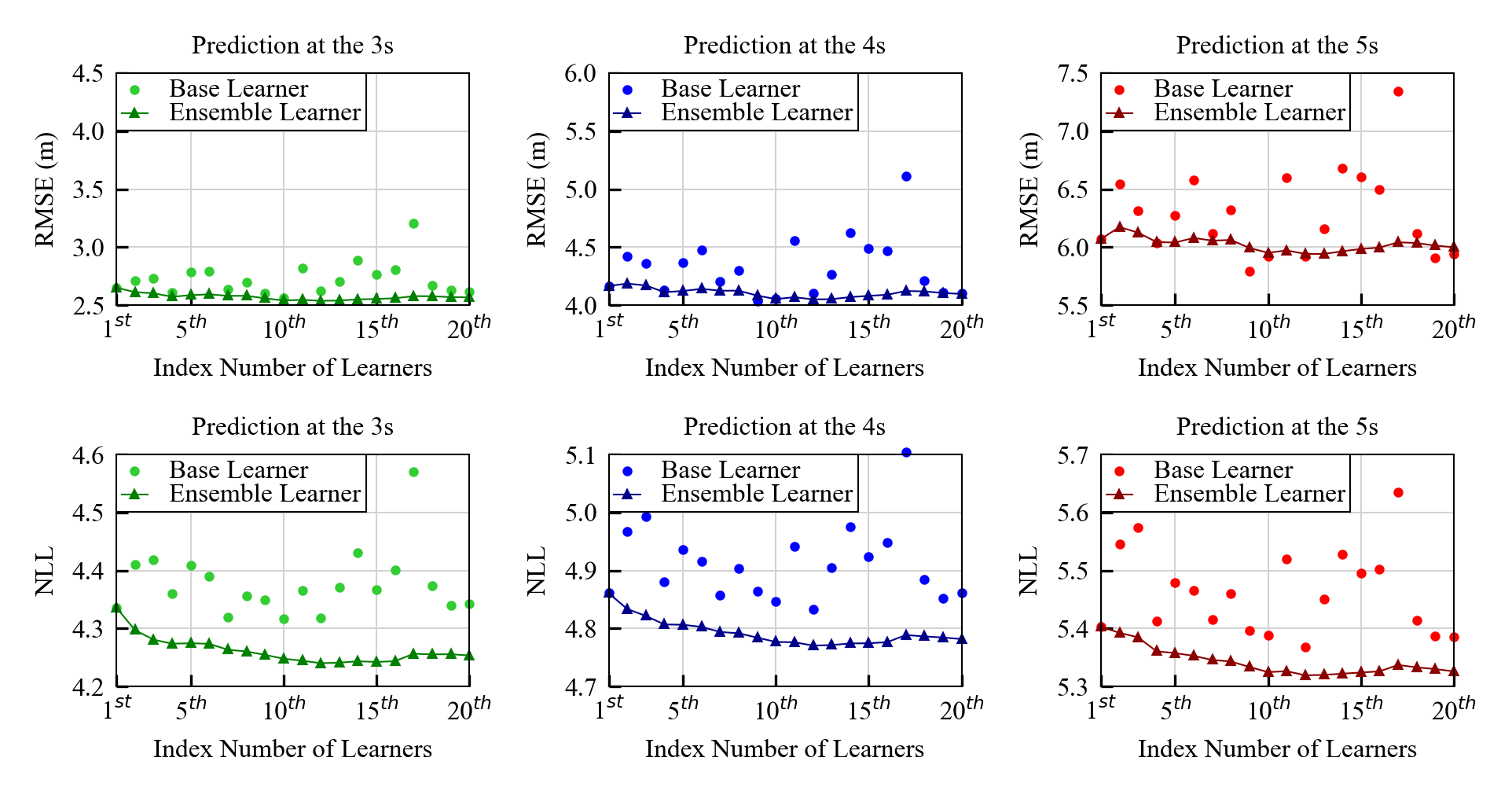}
      \captionsetup{font={small}}
      \caption{Comparisons of RMSE and NLL at the 3s, 4s and 5s in the second experiment. The dot markers represent the performance of base learners, and the triangle markers represent the performance of ensemble learners. In this experiment, the base learners are CS-LSTM-M trained with the  “US-101-0750-0805” segment of the NGSIM dataset, and the ensemble learners are IETP-CS-M. The test set is also split from the processed “US-101-0750-0805” segment.}
      \label{scatters}
 \end{figure*} 

The first experiment is to evaluate the predicting accuracy of IETP. The training and test sets are split from the whole processed NGSIM datasets. The second experiment compares the performance of IETP and the base learners, where training and test sets are split from the “US-101-0750-0805” segment. The third experiment is to evaluate the data efficiency of IETP comparing to several baselines. Each base learner in IETP is trained with 50\% of the training data compared to the first experiment. Then, IETP is tested on the same test set in the first experiment.

\subsection{Dataset and Data-processing} 
The publicly available dataset NGSIM including US-101 and I-80, is used in this study. The dataset consists of real freeway traffic and traffic contexts, which can present interactive scenarios in freeways. In each experiment, features including time frames, coordinates of vehicles are firstly extracted. Then, the dataset is split into a training set and a test set. The test set has a fourth of samples from the dataset. Finally, as described in Section \ref{methods}, 20 sub-training sets are obtained by the Bootstrap sampling based on the whole training set.

\subsection{Evaluating Metrics and Baseline Models}

According to \cite{deo2018convolutional}, the root of mean squared error (RMSE) and negative log-likelihood (NLL) are used to evaluate models in this study. In detail, RMSE is computed by
\begin{equation}
RMSE = \sqrt{\frac{1}{k}\sum_{i=1}^{k} \left ( \mathbf{Y}_{i} -\mathbf{Y}_{i,true}    \right )^2}\label{rmse}
\end{equation} 
and NLL is computed by 
\begin{equation}
NLL =  - \log \left( {\sum\limits_i {{{\mathop{\rm P}\nolimits} _{\bf{G}}}\left( {{\bf{Y}}|{d_i},{\bf{X}}} \right){\mathop{\rm P}\nolimits} \left( {{d_i}|{\bf{X}}} \right)} } \right).\label{nll}
\end{equation} It should be noted that, in (\ref{rmse}), the predicted trajectory ${{{\bf{Y}}_i}}$ is the one under the maneuver with the most probability when using multi-modal outputs. And ${{{\bf{Y}}_{i,{\rm{ true}}}}}$ is the ground truth. It should also be noted that $k$ in (\ref{rmse}) is the number of data samples, and $d_i$ in (\ref{nll}) represents different driving maneuvers.

IETP is compared with following models, which all consider interactions between surrounding vehicles.
\begin{itemize}
\item C-VGMM + VIM: Variational Gaussian mixture models with a Markov random field based on the vehicle interaction module \cite{deo2018would}. It is modified to use maneuvers classes to allow a fair comparison as described in \cite{deo2018convolutional}.
\item GAIL-GRU: Generative adversarial imitation learning model described in \cite{kuefler2017imitating}.
\item M-LSTM: Maneuvers-LSTM model described in \cite{deo2018multi}.
\item S-LSTM: Social Pooling model described in \cite{alahi2016social}.
\item NLS-LSTM: Non-local Social Pooling model described in \cite{messaoud2019non}.
\item CS-LSTM: Convolutional Social Pooling without maneuvers described in \cite{deo2018convolutional}.
\item CS-LSTM-M: Convolutional Social Pooling with maneuvers (CS-LSTM-M) described in \cite{deo2018convolutional}.
\end{itemize}

The system settings of IETP include two types: one uses CS-LSTM as base learners, and the other uses CS-LSTM-M as base learners. These two settings are termed IETP-CS and IETP-CS-M, respectively.

\subsection{Experimental Results and Discussion}

\begin{table*}[ht]
\captionsetup{font={small}}
\caption{COMPARISON OF RMSE OVER THE PREDICTION HORIZON}
\begin{center}
\begin{tabular}{c c c c c c c c c c}
\toprule
Prediction & C-VGMM+ & GAIL-GRU & M-LSTM & S-LSTM & NLS-LSTM & CS-LSTM & CS-LSTM & IETP-CS & IETP-CS \\
horizon (s) & VIM & & & & & -M & & -M (ours) & (ours)\\
\midrule
1 &	0.66&	0.69&	0.58&	0.65&	0.56&	0.62&	0.61&	0.54&	\textbf{0.52} \\
2&	1.56&  	1.51&	1.26&	1.31&	1.22&	1.29&	1.27&	1.22&	\textbf{1.16} \\
3&	2.75&	2.55&	2.12&	2.16&	2.02&	2.13&	2.09&	2.06&	\textbf{1.94} \\
4&	4.24&	3.65&	3.24&	3.25&	3.03&	3.20&	3.10&	3.15&	\textbf{2.95} \\
5&	5.99&	4.71&   4.66&	4.55&	4.30&	4.52&	4.37&	4.52&	\textbf{4.24} \\
\bottomrule
\label{results_rmse}
\end{tabular}
\end{center}
\end{table*}

\subsubsection{Experimental Results}
Table \ref{results_rmse} shows the comparisons of RMSE over the prediction horizon in the first experiment. The RMSE values of baselines are results from \cite{deo2018convolutional, messaoud2019non}. Results of IETP-CS-M and IETP-CS are average results of all ensemble learners. Experimental results show that IETP-CS has the lowest RMSE over the prediction horizon compared to all models. Results also show that IETP-CS and IETP-CS-M decrease the RMSE compared to the individual CS-LSTM and CS-LSTM-M. Fig. \ref{scatters} shows the detailed comparisons of RMSE and NLL in the second experiment. The dot markers in Fig. \ref{scatters} represent the performance of base learners (CS-LSTM-M), and the triangle markers with lines represent the performance of ensemble learners (IETP-CS-M). It should be explained that the $n^{th}$ base learner is the individual base learner trained on the $n^{th}$ sampling set. And the $n^{th}$ ensemble learner is built by assembling $n$ base learner (from the $1^{st}$ to the $n^{th}$). It can be found that ensemble learners have lower RMSE and NLL most of the time. These results show that IETP improves the predicting accuracy of base learners.
   \begin{figure}[tp]
      \centering
      \includegraphics[scale=1.0]{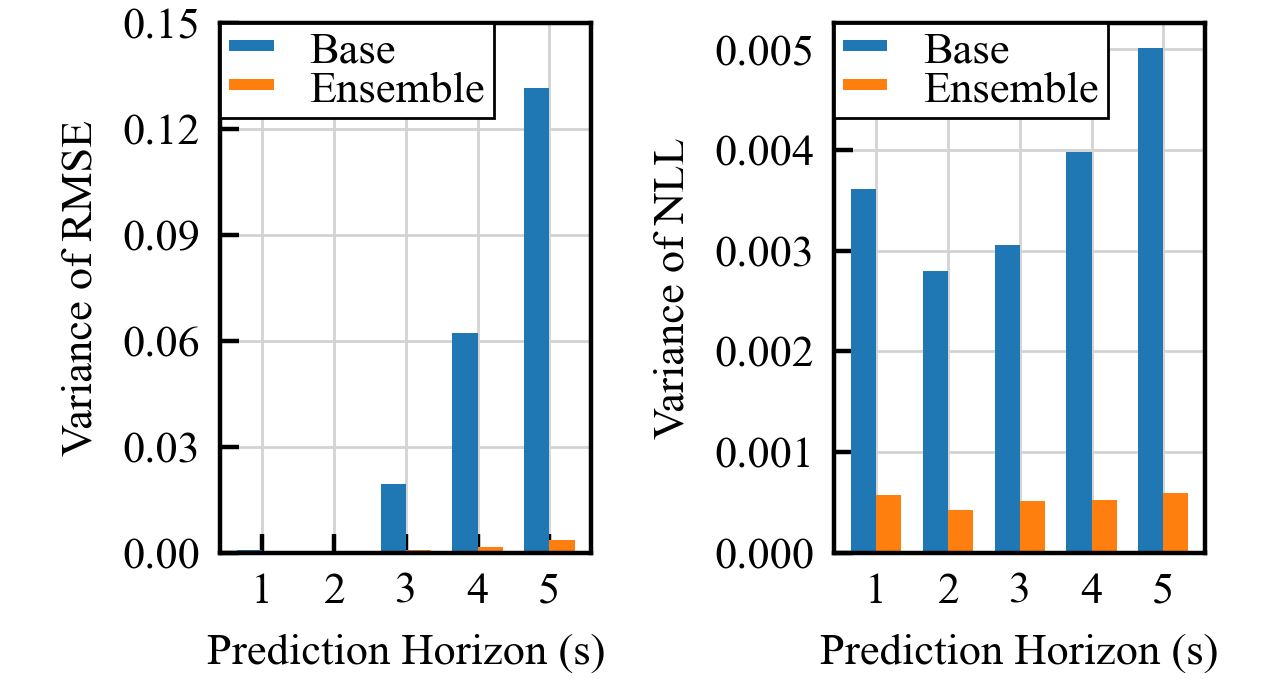}
      \captionsetup{font={small}}
      \caption{Comparisons of variances of RMSE (left) and NLL (right) on all base learners and all ensemble learners.}
      \label{var}
   \end{figure}
  \begin{figure}[tp]
  \centering
  \includegraphics[scale=1.0]{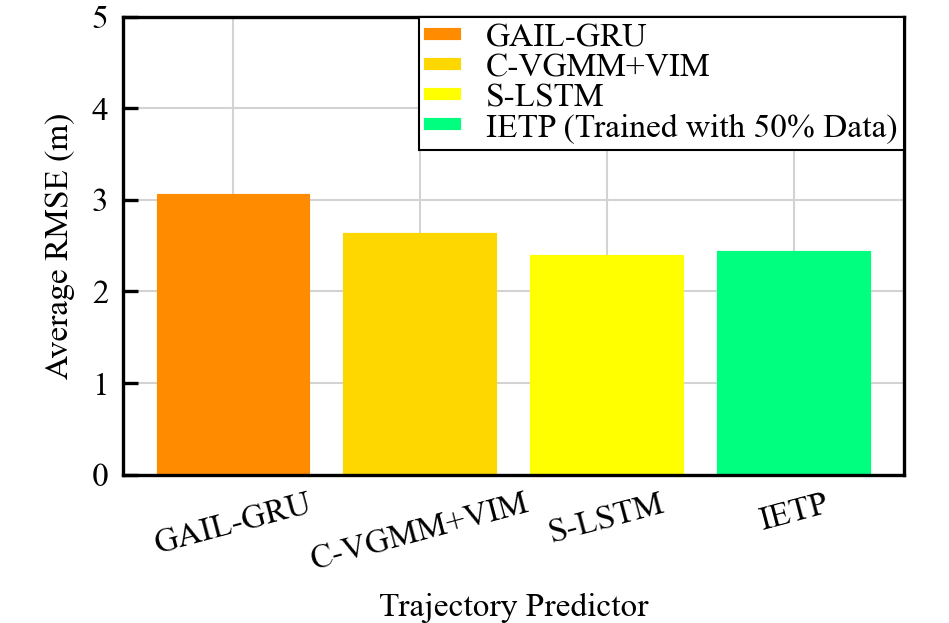}
  \captionsetup{font={small}}
  \caption{Average RMSE over prediction horizon. The IETP refers to the IETP-CS trained with 50\% data of the whole NGSIM training set.}
  \label{light}
\end{figure}
Fig. \ref{var} shows the variance of the RMSE and NLL from 20 base learners (CS-LSTM-M) and 20 ensemble learners (IETP-CS-M) in the second experiment. As shown in Fig. \ref{var}, base learners have a higher variance of the RMSE and NLL than ensemble learners. Moreover, the RMSE and NLL are decreased significantly by applying the ensemble learning approach, especially when the prediction horizon extends further. For example, at the 4s, the variance of RMSE and NLL are decreased by 98\% and 87\%. And at the 5s, RMSE and NLL are reduced by 97\% and 88\%, respectively. These results can also be found intuitively in Fig. \ref{scatters}. As shown in Fig. \ref{scatters}, the performance of base learners varies wider than the ensemble learners among 20 groups of testing. Moreover, the gap between the best and the worst performance of base learners can be enormous. For example, at the 5s, the gap between the highest and the lowest RMSE of base learners is 1.56 m. In comparison, the largest gap of ensemble learners is 0.23 m. These results show that IETP has a more stable performance than base learners.

Fig. \ref{light} shows the comparison of average RMSE over the prediction horizon in the third experiment. From Fig. \ref{light}, IETP-CS trained with 50\% data outperforms C-VGMM+VIM and GAIL-GRU. Besides, the average RMSE values of S-LSTM and IETP-CS are 2.38m and 2.42m, which are very close.
    \begin{figure}[tp]
      \centering
      \includegraphics[scale=1.0]{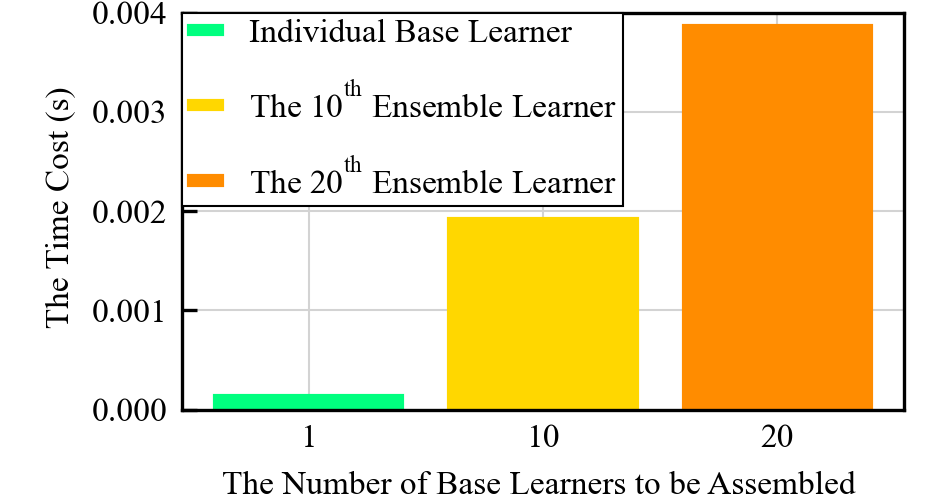}
      \captionsetup{font={small}}
      \caption{The time cost to predict trajecotries on a sample: base learner (left), the 10$^{th}$ ensemble learner (middle), and the 20$^{th}$ ensemble learner (right).}
      \label{timecost}
   \end{figure}

\subsubsection{ Comparative Study in the Time-cost}
Compared to the individual model, more models need to be loaded in IETP. As a result, it takes more time for ensemble learners to predict trajectories. The time cost of models to predict trajectories on each sample is recorded in this study. Recordings show that the relationship between the time cost and the number of assembled base learners is almost linear. Fig. \ref{timecost} shows the time cost of models to predict trajectories on each sample. However, we found that using more base learners to build ensemble learners does not significantly improve the performance. For example, from Fig. \ref{scatters}, the performance of the 10$^{th}$ ensemble learner and the 20$^{th}$ ensemble learner are close to each other. In contrast, the 10$^{th}$ ensemble learner has a lower time cost. 

The best-performed ensemble learner is different among experiments. This paper may not propose a specific number of base learners to be used. Nevertheless, even the highest time cost, i.e., the time cost of the 20$^{th}$ ensemble learner in experiments, is approximately four milliseconds. Since the timescale for vehicles to change lanes or make a brake action in interactive scenarios is about a few seconds \cite{xing2020ensemble}, the time cost of IETP is acceptable.

\section{Conclusions}
\label{conclu}
In this paper, an ensemble learning-based framework is proposed for vehicle trajectory prediction in interactive scenarios. The proposed framework is built by assembling interaction-aware trajectory predictors through the Bootstrap aggregating. Two ensemble strategies are applied to solve the driving maneuvers classification and trajectory prediction tasks. By adopting Convolutional Social Pooling models as base learners, three experiments are conducted on the NGSIM dataset to evaluate the framework. 

Firstly, experimental results show that the proposed framework improves the predicting accuracy compared to base learners. Secondly, experimental results show that the performance variance can be significantly decreased by applying the proposed framework. Since the variance measures the dispersion of the data, it shows that individual models can have a more stable performance by applying the proposed framework. Moreover, the proposed framework with less training data outperforms the GAIL-GRU \cite{kuefler2017imitating} and C-VGMM+VIM \cite{deo2018would} models while reaching a similar performance of the S-LSTM \cite{alahi2016social}. It indicates that the proposed framework is data-efficient.  

In this study, specific base learners in our ensemble learning framework are Convolutional Social Pooling models. However, the proposed ensemble learning framework provides a general ensemble learning-based approach to predict trajectories in interactive scenarios. Since the characteristics and number of base learners are critical to the performance of ensemble learners, selecting different types and numbers of models as base learners will be considered and discussed in future works.




\bibliographystyle{ieeetr}
\bibliography{ref.bib}

\begin{thebibliography}{10}

\bibitem{li2022personalized}
Z.~Li, J.~Gong, C.~Lu, and J.~Li, ``Personalized driver braking behavior
  modelling in the car-following scenario: An importance weight-based transfer
  learning approach,'' {\em IEEE Transactions on Industrial Electronics}, 2022.

\bibitem{lu2019transfer}
C.~Lu, F.~Hu, D.~Cao, J.~Gong, Y.~Xing, and Z.~Li, ``Transfer learning for
  driver model adaptation in lane-changing scenarios using manifold
  alignment,'' {\em IEEE transactions on intelligent transportation systems},
  vol.~21, no.~8, pp.~3281--3293, 2019.

\bibitem{lu2019virtual}
C.~Lu, F.~Hu, D.~Cao, J.~Gong, Y.~Xing, and Z.~Li, ``Virtual-to-real knowledge
  transfer for driving behavior recognition: Framework and a case study,'' {\em
  IEEE Transactions on Vehicular Technology}, vol.~68, no.~7, pp.~6391--6402,
  2019.

\bibitem{deo2018would}
N.~Deo, A.~Rangesh, and M.~M. Trivedi, ``How would surround vehicles move? a
  unified framework for maneuver classification and motion prediction,'' {\em
  IEEE Transactions on Intelligent Vehicles}, vol.~3, no.~2, pp.~129--140,
  2018.

\bibitem{li2020importance}
Z.~Li, J.~Gong, C.~Lu, and J.~Xi, ``Importance weighted gaussian process
  regression for transferable driver behaviour learning in the lane change
  scenario,'' {\em IEEE Transactions on Vehicular Technology}, vol.~69, no.~11,
  pp.~12497--12509, 2020.

\bibitem{ess2010object}
A.~Ess, K.~Schindler, B.~Leibe, and L.~Van~Gool, ``Object detection and
  tracking for autonomous navigation in dynamic environments,'' {\em The
  International Journal of Robotics Research}, vol.~29, no.~14, pp.~1707--1725,
  2010.

\bibitem{luber2010people}
M.~Luber, J.~A. Stork, G.~D. Tipaldi, and K.~O. Arras, ``People tracking with
  human motion predictions from social forces,'' in {\em 2010 IEEE
  International Conference on Robotics and Automation}, pp.~464--469, IEEE,
  2010.

\bibitem{liu2021survey}
J.~Liu, X.~Mao, Y.~Fang, D.~Zhu, and M.~Q.-H. Meng, ``A survey on deep-learning
  approaches for vehicle trajectory prediction in autonomous driving,'' {\em
  arXiv preprint arXiv:2110.10436}, 2021.

\bibitem{alahi2016social}
A.~Alahi, K.~Goel, V.~Ramanathan, A.~Robicquet, L.~Fei-Fei, and S.~Savarese,
  ``Social lstm: Human trajectory prediction in crowded spaces,'' in {\em
  Proceedings of the IEEE conference on computer vision and pattern
  recognition}, pp.~961--971, 2016.

\bibitem{deo2018convolutional}
N.~Deo and M.~M. Trivedi, ``Convolutional social pooling for vehicle trajectory
  prediction,'' in {\em Proceedings of the IEEE Conference on Computer Vision
  and Pattern Recognition Workshops}, pp.~1468--1476, 2018.

\bibitem{xu2018collision}
K.~Xu, Z.~Qin, G.~Wang, K.~Huang, S.~Ye, and H.~Zhang, ``Collision-free lstm
  for human trajectory prediction,'' in {\em International Conference on
  Multimedia Modeling}, pp.~106--116, Springer, 2018.

\bibitem{zhang2019sr}
P.~Zhang, W.~Ouyang, P.~Zhang, J.~Xue, and N.~Zheng, ``Sr-lstm: State
  refinement for lstm towards pedestrian trajectory prediction,'' in {\em
  Proceedings of the IEEE/CVF Conference on Computer Vision and Pattern
  Recognition}, pp.~12085--12094, 2019.

\bibitem{zhao2020spatial}
X.~Zhao, Y.~Chen, J.~Guo, and D.~Zhao, ``A spatial-temporal attention model for
  human trajectory prediction.,'' {\em IEEE CAA J. Autom. Sinica}, vol.~7,
  no.~4, pp.~965--974, 2020.

\bibitem{peng2021sra}
Y.~Peng, G.~Zhang, J.~Shi, B.~Xu, and L.~Zheng, ``Sra-lstm: Social relationship
  attention lstm for human trajectory prediction,'' {\em arXiv preprint
  arXiv:2103.17045}, 2021.

\bibitem{gupta2018social}
A.~Gupta, J.~Johnson, L.~Fei-Fei, S.~Savarese, and A.~Alahi, ``Social gan:
  Socially acceptable trajectories with generative adversarial networks,'' in
  {\em Proceedings of the IEEE Conference on Computer Vision and Pattern
  Recognition}, pp.~2255--2264, 2018.

\bibitem{kosaraju2019social}
V.~Kosaraju, A.~Sadeghian, R.~Mart{\'\i}n-Mart{\'\i}n, I.~Reid, S.~H.
  Rezatofighi, and S.~Savarese, ``Social-bigat: Multimodal trajectory
  forecasting using bicycle-gan and graph attention networks,'' {\em arXiv
  preprint arXiv:1907.03395}, 2019.

\bibitem{sun2020recursive}
J.~Sun, Q.~Jiang, and C.~Lu, ``Recursive social behavior graph for trajectory
  prediction,'' in {\em Proceedings of the IEEE/CVF Conference on Computer
  Vision and Pattern Recognition}, pp.~660--669, 2020.

\bibitem{zhao2020gisnet}
Z.~Zhao, H.~Fang, Z.~Jin, and Q.~Qiu, ``Gisnet: Graph-based information sharing
  network for vehicle trajectory prediction,'' in {\em 2020 International Joint
  Conference on Neural Networks (IJCNN)}, pp.~1--7, IEEE, 2020.

\bibitem{wang2021graphtcn}
C.~Wang, S.~Cai, and G.~Tan, ``Graphtcn: Spatio-temporal interaction modeling
  for human trajectory prediction,'' in {\em Proceedings of the IEEE/CVF Winter
  Conference on Applications of Computer Vision}, pp.~3450--3459, 2021.

\bibitem{li2021hierarchical}
Z.~Li, C.~Lu, Y.~Yi, and J.~Gong, ``A hierarchical framework for interactive
  behaviour prediction of heterogeneous traffic participants based on graph
  neural network,'' {\em IEEE Transactions on Intelligent Transportation
  Systems}, 2021.

\bibitem{li2021interactive}
Z.~Li, J.~Gong, C.~Lu, and Y.~Yi, ``Interactive behavior prediction for
  heterogeneous traffic participants in the urban road: A
  graph-neural-network-based multitask learning framework,'' {\em IEEE/ASME
  Transactions on Mechatronics}, vol.~26, no.~3, pp.~1339--1349, 2021.

\bibitem{xing2020ensemble}
Y.~Xing, C.~Lv, H.~Wang, D.~Cao, and E.~Velenis, ``An ensemble deep learning
  approach for driver lane change intention inference,'' {\em Transportation
  Research Part C: Emerging Technologies}, vol.~115, p.~102615, 2020.

\bibitem{sagi2018ensemble}
O.~Sagi and L.~Rokach, ``Ensemble learning: A survey,'' {\em Wiley
  Interdisciplinary Reviews: Data Mining and Knowledge Discovery}, vol.~8,
  no.~4, p.~e1249, 2018.

\bibitem{dong2020survey}
X.~Dong, Z.~Yu, W.~Cao, Y.~Shi, and Q.~Ma, ``A survey on ensemble learning,''
  {\em Frontiers of Computer Science}, vol.~14, no.~2, pp.~241--258, 2020.

\bibitem{dietterich2002ensemble}
T.~G. Dietterich {\em et~al.}, ``Ensemble learning,'' {\em The handbook of
  brain theory and neural networks}, vol.~2, no.~1, pp.~110--125, 2002.

\bibitem{ngsimdata}
B.~Coifman and L.~Li, ``A critical evaluation of the next generation simulation
  (ngsim) vehicle trajectory dataset,'' {\em Transportation Research Part B:
  Methodological}, vol.~105, pp.~362--377, 2017.

\bibitem{efron1994introduction}
B.~Efron and R.~J. Tibshirani, {\em An introduction to the bootstrap}.
\newblock CRC press, 1994.

\bibitem{kingma2014adam}
D.~P. Kingma and J.~Ba, ``Adam: A method for stochastic optimization,'' {\em
  arXiv preprint arXiv:1412.6980}, 2014.

\bibitem{kuefler2017imitating}
A.~Kuefler, J.~Morton, T.~Wheeler, and M.~Kochenderfer, ``Imitating driver
  behavior with generative adversarial networks,'' in {\em 2017 IEEE
  Intelligent Vehicles Symposium (IV)}, pp.~204--211, IEEE, 2017.

\bibitem{deo2018multi}
N.~Deo and M.~M. Trivedi, ``Multi-modal trajectory prediction of surrounding
  vehicles with maneuver based lstms,'' in {\em 2018 IEEE Intelligent Vehicles
  Symposium (IV)}, pp.~1179--1184, IEEE, 2018.

\bibitem{messaoud2019non}
K.~Messaoud, I.~Yahiaoui, A.~Verroust-Blondet, and F.~Nashashibi, ``Non-local
  social pooling for vehicle trajectory prediction,'' in {\em 2019 IEEE
  Intelligent Vehicles Symposium (IV)}, pp.~975--980, IEEE, 2019.

\end{thebibliography}

\end{document}